\documentclass{article}
\usepackage{spconf,amsmath,graphicx,amsfonts}

\usepackage{enumitem}
\setlist{nosep, leftmargin=14pt}

\usepackage{mwe} % to get dummy images
\usepackage{xcolor}
\usepackage{subcaption}
\newcommand{\vect}[1]{\boldsymbol{#1}}

\title{Multimodal Learning To Improve Cardiac Late Mechanical \\ Activation Detection From Cine MR Images}
\name{
\parbox{\linewidth}{
\centering
Jiarui Xing$^{ a}$ \qquad
Nian Wu$^{ a}$ \qquad\\
Kenneth C. Bilchick$^{ d}$ \qquad
Frederick H. Epstein$^{ c}$ \qquad
Miaomiao Zhang$^{ a, b}$
}}

\address{
\parbox{\linewidth}{
\centering
$^{a}$ Department of Electrical and Computer Engineering, University of Virginia, USA \\
$^{b}$ Department of Computer Science, University of Virginia, USA  \\
$^{c}$ Department of Biomedical Engineering, University of Virginia Health System, USA\\
$^{d}$ School of Medicine, University of Virginia Health System, USA%\\
}}
\begin{document}
%\ninept
%
\maketitle
\begin{abstract}
This paper presents a multimodal deep learning framework that utilizes advanced image techniques to improve the performance of clinical analysis heavily dependent on routinely acquired standard images. More specifically, we develop a joint learning network that for the first time leverages the accuracy and reproducibility of myocardial strains obtained from Displacement Encoding with Stimulated Echo (DENSE) to guide the analysis of cine cardiac magnetic resonance (CMR) imaging  in late mechanical activation (LMA) detection. An image registration network is utilized to acquire the knowledge of cardiac motions, an important feature estimator of strain values, from standard cine CMRs. Our framework consists of two major components: (i) a DENSE-supervised strain network leveraging latent motion features learned from a registration network to predict myocardial strains; and (ii) a LMA network taking advantage of the predicted strain for effective LMA detection. Experimental results show that our proposed work substantially improves the performance of strain analysis and LMA detection from cine CMR images, aligning more closely with the achievements of DENSE.

\end{abstract}

%
%\begin{keywords}
%Multimodal learning, joint optimization, late mechanical activation detection
%\end{keywords}
%
\section{Introduction}
\label{sec:intro}
Myocardial strain has demonstrated its significance in identifying LMA regions for an optimized pacing site for cardiac resynchronization therapy (CRT)~\cite{bilchick2020cmr,budge2012mr}. The quantification of myocardial strains can be achieved through various specialized imaging techniques that offer information of ventricular deformation patterns and cardiac motion abnormalities from MR images. Commonly used methods include MR tagging~\cite{axel1989heart}, cine SSFP with feature tracking (FT)~\cite{tee2013imaging,morales2021deepstrain,qiao2020temporally,qin2018joint}, and cine DENSE~\cite{kim2004myocardial}, with DENSE standing out for its high accuracy in capturing myocardial deformations~\cite{kar2014validation}. Despite the advantages of DENSE, its widespread clinical use is hindered by limited accessibility, primarily due to the high-cost facilities and specialized expertise required for image collection and analysis. Many clinical centers often opt for cine FT. However, the accuracy of FT is compromised by inherent limitations in image quality, including low spatial and temporal resolution. Additionally, these registration-based tracking algorithms focus solely on motions along contours~\cite{young2012generalized}. 
%Accurate strain analysis can be computed from various specialized imaging techniques for cardiac motion detection, such as myocardial tagging~\cite{axel1989heart}, displacement encoding with stimulated echoes (DENSE)\cite{kim2004myocardial}, and strain-encoded imaging\cite{osman2001imaging}. Among these, DENSE stands out for its high accuracy in capturing myocardial displacement~\cite{kar2014validation}. However, these techniques often suffer from limited clinical accessibility due to the specialized expertise needed for data collection and analysis~\cite{frija2021improve}. An alternative approach is to use registration or feature tracking on bSSFP cine MR images, which are more straightforward to acquire and analyze~\cite{tee2013imaging}. There have been many existing methods trying to utilize the motion captured from cine data~\cite{morales2021deepstrain,qiao2020temporally,qin2018joint}. However, the accuracy is compromised by the inherent limitation of image quality and resolution of the cine data~\cite{young2012generalized}, which drives ongoing research efforts to make improvement.

Recent research has explored the application of deep learning to enhance the accuracy of predicting myocardial motion from cine images, guided by the supervision of DENSE~\cite{wang2023strainnet}. In this study, the authors employed a neural network to capture the intricate relationship between a time sequence of left ventricular (LV) myocardium segmented from DENSE, and the corresponding encoded displacement fields. By assuming a minimal domain gap between cine and DENSE segmentations in predicting displacement fields, the researchers directly evaluated the trained model on cine input.

Inspired by~\cite{wang2023strainnet}, this paper introduces  a multimodal deep learning method that for the first time leverages DENSE to guide the analysis of cine CMRs for an improved LMA detection. Using DENSE strain as ground truth data, we develop an end-to-end joint learning framework that predicts LMA regions (measured by the onset of circumferential shortening (TOS) of segmental myocardial regions~\cite{wyman1999mapping}) from cine images. Our framework includes two main components: (i) a registration-based strain network to predict the myocardium strain using the learned latent motion/deformation features from cine images, and (ii) a LMA network to predict TOS based on the learned strains. These subnetworks are simultaneously trained to mutually benefit each other, resulting in improved overall performance. 

To the best of our knowledge, our method is the first to leverage machine learning to improve LMA detection from cine images, guided by DENSE. This opens promising research venues for transferring knowledge from advanced strain imaging to routinely acquired CMR data. Additionally, our method increases the accessibility to DENSE, particularly in under-resourced regions and populations. Our experimental results demonstrate a substantial improvement in LMA detection accuracy compared to exsiting approaches. Future work will involve meticulous validation of model generalizability as additional patient data becomes available. 

%In this paper, we present a novel multimodal joint learning approach that effectively avoiding the negative effect from the domain difference by using cine data as input and DENSE as a training guide. Our framework includes an image registration network to capture cardiac motion from cine data, and a strain prediction network that takes the latent feature of the registration network containing rich cardiac motion information, to predict strain. The accurate strain data from DENSE serve to supervise the network prediction. Additionally, we incorporate an LMA network for LMA detection. During training, these networks are simultaneously trained, enabling the mutual guidance between the strain prediction and LMA detection, leading to improved performance of both tasks. Our contributions are twofold: Firstly, to the best of our knowledge, this is the pioneer work that employs DENSE as a guide for predicting accurate strain from cine data for the LMA detection task, effectively bridging the domain gap between DENSE and cine data seen in previous pre-training approach. Secondly, we validate our approach on LMA region estimation. Experimental results demonstrate a significant enhancement in LMA detection accuracy, highlighting the effectiveness of our proposed approach.
\begin{figure*}[!htb]
    \centering    
    \includegraphics[width=\linewidth]{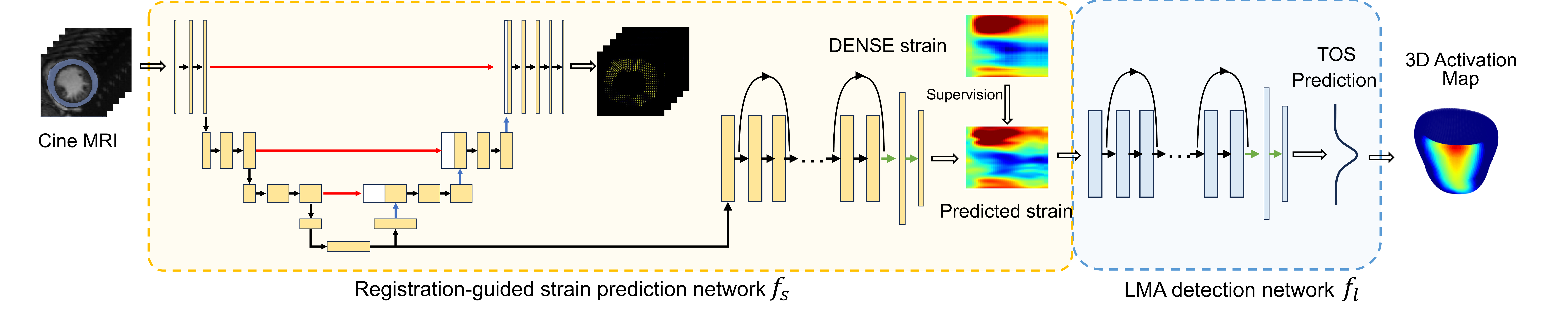}
    \caption{The multimodal joint learning framework with registration-guided strain prediction and LMA detection networks.}
    \label{fig:framework}
    \vspace{-1.2em}
\end{figure*}
\section{Methodology}
This section presents our joint learning framework of two submodules, including a registration-guided strain prediction network guided by DENSE and a LMA detection network (as illustrated in Fig.~\ref{fig:framework}). Before introducing our model, we will first briefly review CMR myocardial strain analysis. 
%{\color{blue}In this section, we first briefly review the concept of myocardial strain analysis from CMR images, and then present our joint learning framework, which contains a registration-guided strain prediction network guided by DENSE and a LMA detection network (as illustrated in Fig.~\ref{fig:framework})}

\begin{figure}[ht!]
\centering
\begin{subfigure}{0.29\linewidth}
  \centering
  \includegraphics[width=\textwidth]{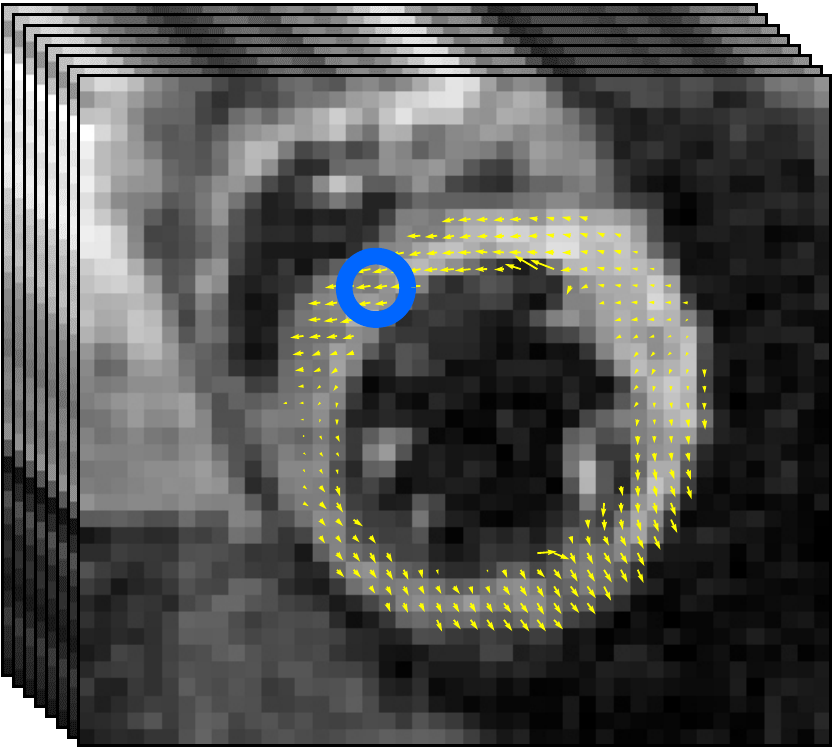}
    \caption{}
  \label{fig:disp}
\end{subfigure} %
% \hspace{.7cm}
\begin{subfigure}{.29\linewidth}
  \centering
  \includegraphics[width=\textwidth]{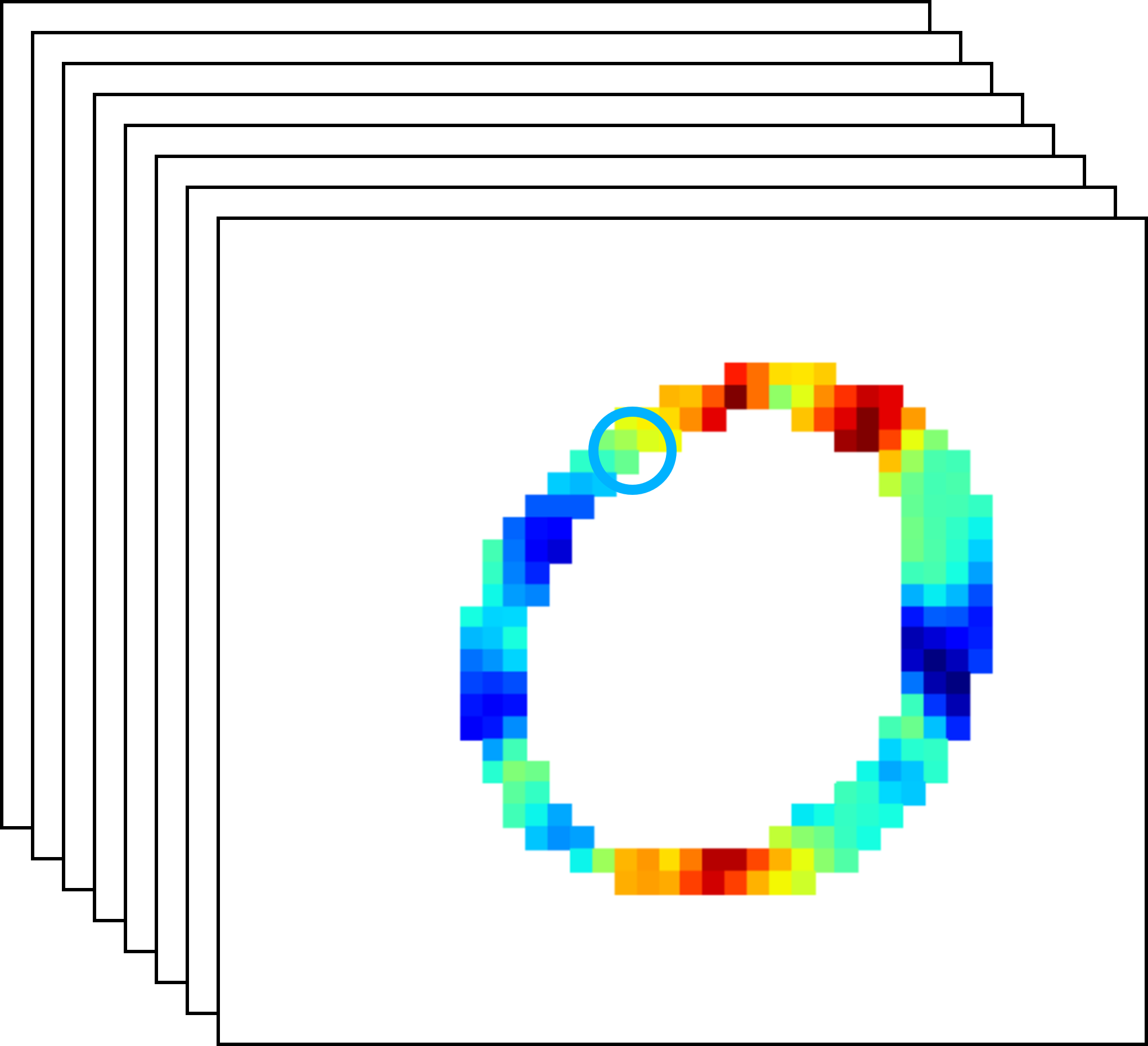}
  \caption{}
  \label{fig:sectors}
\end{subfigure} 
% \hspace{1cm}
\begin{subfigure}{.37\linewidth}
  \centering
  \includegraphics[width=\textwidth]{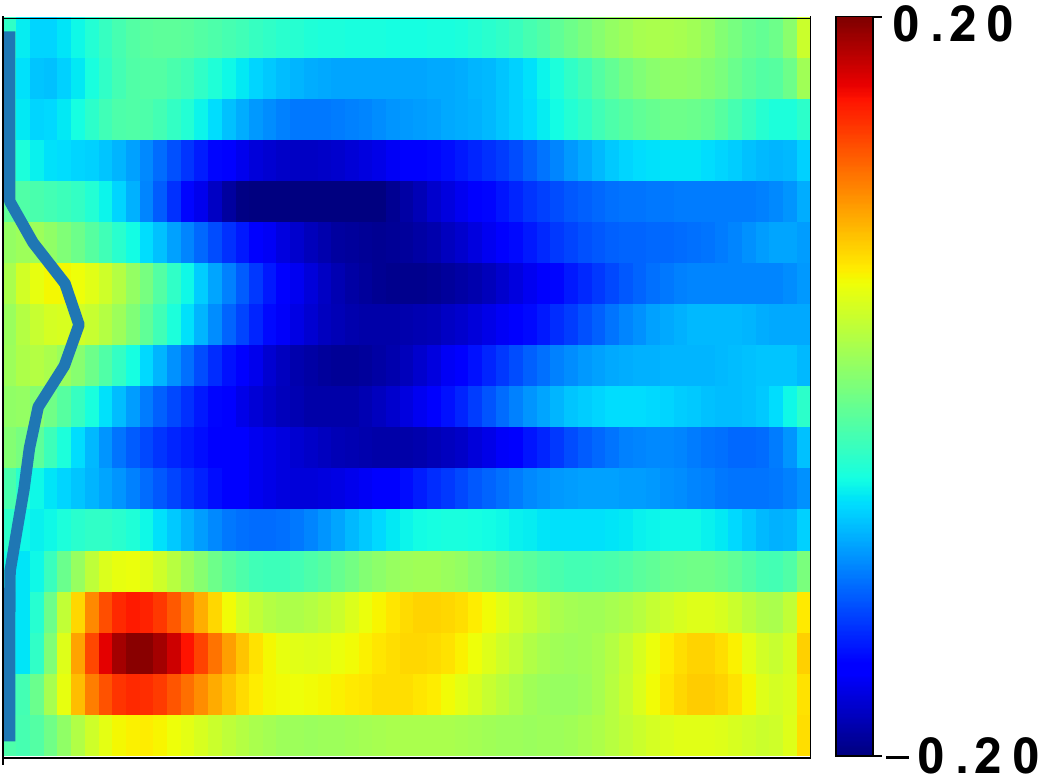}
    \caption{}
  \label{fig:strainMat_TOS}
\end{subfigure} %
% \begin{subfigure}{.49\linewidth}
%   \centering
%   \includegraphics[width=\linewidth]{Imgs/Background/strainImg-SET01CT01MMid1F15.pdf}
%     \caption{}
%   \label{fig:strainImg_1}
% \end{subfigure}
\caption{Example of (a) temporal CMRs overlaid with displacement fields; (b) LV strain (contraction/stretching in blue/red; the blue circle shows the sampling starting location); and (c) 2D strain matrix and its corresponding TOS curve.}
\vspace{-2em}
\label{fig:strain Analysis}
\end{figure}

\noindent \textbf{CMR strain analysis.}
%\vspace{-0.5em}
% Initial submission
% Consider a time-sequence of CMR images with $T$ time frames (see Fig.~\ref{fig:strain Analysis}(a)). For each time frame, we compute a $N$-dimensional strain vector based on the displacement fields from a number of $N$ myocardial sectors, beginning from the the middle of the intersection points of the left and right ventricle and following counter-clockwise order. A $N \times T$ strain matrix that includes information from all time frames is built by concatenating the strain vectors across time (see Fig.~\ref{fig:strain Analysis}(b)). A TOS curve labeled from the 2D strain matrix is shown in Fig.~\ref{fig:strain Analysis}(c).
% Jerry edited - 240217
Consider a time-sequence of CMR images with $T$ time frames (see Fig.~\ref{fig:strain Analysis}(a)). For each time frame, we compute the circumferential strain along the myocardium based on the displacement fields, and sample the strain values into $N$-dimensional strain vector from a number of $N$ evenly divided myocardial sectors, beginning from the the middle of the intersection points 
% of the left and right ventricle 
and following counter-clockwise order (see Fig.~\ref{fig:strain Analysis}(b)). A $N \times T$ strain matrix containing information from all time frames is built by concatenating the strain vectors across time. A TOS curve labeled from the 2D strain matrix is shown in Fig.~\ref{fig:strain Analysis}(c). Here, each TOS value represents the start time of contraction in the corresponding sector, with higher values indicating more severe LMA due to delayed contraction~\cite{bilchick2020cmr}.

%In our framework, three networks are trained simultaneous: (1) an image registration network to extract the motion information from the cine image sequence, (2) a strain prediction network to predict the myocardium strain from the latent space of the registration work, and (3) a LMA task network to predict the the onset of circumferential shortening (TOS) of the myocardium sectors, as TOS has been proven to have a close linear relationship with electrical activation time and the ability to detect LMA regions~\cite{wyman1999mapping}. 
% With the guidance of advanced DENSE data and LMA task during the training phase, the models become capable of making precise LMA task predictions using highly accessible cine MR image sequences.

%\vspace{-1em}
\subsection{Our Multimodal Learning Network}
%\vspace{-0.5em}
% \noindent
% \textbf{Task formulation}. 
% As the displacement captured from DENSE and cine data has significantly difference, to mitigate this domain gap we choose to follow the previous studies~\cite{xing2023multitask} to predict accurate circumferential myocardium strain matrix instead, which is a robust and effectiveness spatial temporal representation of cardiac motion~\cite{budge2012mr}. To improve the prediction performance and avoid overfitting, we adopt an image registration network to predict the Lagrangian displacement from cine image sequence, and pass its latent feature containing rich motion information to a strain prediction network. 
%$f_{\text{reg}}$ is trained to generate the initial velocity $v_0$ given a pair of source and target images. Then the final smooth displacement field is calculated by numerical integrating the Euler-Poincaré differential equation (EPDiff)~\cite{miller2006} given the initial velocity. 

% \noindent\textbf{Network architecture and loss function}. 
\paragraph*{Registration-based strain network.} Given a time-sequence of cine CMRs, $\{I_t\}$, where $t \in [1, \cdots, T]$, we employ a registration network to first learn the latent features of cardiac motions, represented by Lagrangian displacement fields $\{\phi_t\}$, from images. Such latent features, denoted as $z$, are directly utilized to predict strains with the supervision of DENSE strain data. We employ a UNet architecture backbone~\cite{ronneberger2015u} for our registration network, and a ResNet network for the strain prediction~\cite{he2016deep,xing2023multitask}. Analogues to~\cite{ramachandran2015singular}, we apply a low-rank singular value decomposition to the predicted strain matrix for smoothness constraints.
% Other architectures, such as Unet++~\cite{zhou2018unet++} or transUNet~\cite{chen2021transunet}, can be easily applied. 
%Analogous to the previous work~\cite{xing2023multitask}, we adopt a ResNet~\cite{he2016deep}-based 3D convolutional network for strain prediction. 
% Note that we apply a low-rank singular value decomposition to the predicted strain matrix for smoothness constraints~\cite{ramachandran2015singular}. 

Defining DENSE strain as $S$, and $\theta_r, \theta_s$ as the registration network, and strain network parameters respectively, we can now formulate the loss function of our registration-based strain network as
\begin{align}
    l_{\text{strain}} = 
    &\sum_{t=1}^{T}{\left[\frac{1}{2\sigma^{2}}\| I_1 \circ \phi_t(\theta_r) - I_{t} \|_2^2
            + \text{Reg}(\phi_t(\theta_r))\right]} \nonumber\\
    & + \alpha \|f_s(z; \theta_s) - S \|_2^2
    + \lambda \|\theta_r\|_2^2 
    + \mu \|\theta_s\|_2^2,    
    \label{eq:strain}
\end{align}
where $\circ$ represents interpolation, and $(\sigma, \alpha, \lambda, \mu)$ are positive weighting parameters. The $\text{Reg}(\cdot)$ is a regularization term that encourage the smoothness of the predicted displacement field, $\phi_t$. We adopt the regularization term used in large deformation diffeomorphic metric mapping~\cite{beg2005computing}. 
%Other commonly used regularity terms include those used in stationary velocity field~\cite{arsigny2006log}, and optical flow~\cite{lucas1981iterative}.} \\

% \begin{figure*}[!hb]
%     \centering    
%     \includegraphics[width=\linewidth]{Figures/framework3.pdf}
%     \caption{The framework of our proposed multimodal joint learning framework. The cardiac motion capturing model is trained with guidance from the accurate DENSE data and downstream LMA task simultaneously. In the deploy stage, it generates better motion prediction from accessible cine MR image sequence, and optimize the LMA task performance.}
%     \label{fig:framework}
%     \vspace{-1.2em}
% \end{figure*}

%\vspace{-1em}
\noindent \textbf{LMA regression network to predict TOS.} Analogous to~\cite{xing2023multitask}, we develop a LMA regression network to predict the TOS (a $N$-dimensional vector). Given the predicted strain matrix from Eq.~\eqref{eq:strain}, we utilize a mean-squared-error of predicted TOS and manually labeled ground truth TOS, denoted as $\vect{y}$, for network loss, i.e., 
%across \( N_s \) evenly divided sectors along the myocardium.
% The predicted displacement fields are utilized as inputs for the downstream LMA task network $f_{\text{LMA}}$. This network performs either of the two tasks: time to onset of circumferential shortening (TOS) regression or LMA sector classification. The TOS values of a sector of the myocardium represent the its time to start contraction. Sectors with delayed contraction, indicated by higher TOS values, suggest more severe LMA. The TOS regression task is aimed at predicting TOS in $N_s$ sectors along the myocardium. For the LMA sector classification task, sectors are labeled as LMA if their TOS values exceed a specified threshold, and the aim is to identify each sector as either LMA or non-LMA. 
%The predicted strain matrix feed into the downstream LMA task network \( f_{\text{LMA}} \), which performs time to onset of circumferential shortening (TOS) regression task. TOS values represent the start time of contraction in myocardial sectors, with higher values indicating more severe LMA due to delayed contraction. The network aims to predict TOS across \( N_s \) evenly divided sectors along the myocardium.
\begin{equation}
    l_{\text{TOS}} = \beta \|f_{l}(f_s(z; \theta_s);\theta_l) - \vect{y}\|_2^2 + \gamma \| \theta_{l} \|_2^2 ,
    \label{eq:LMA}
\end{equation}
where $\theta_{l}$ represents network parameters, with $\beta$ and $\gamma$ being the weighting parameters. 
%Note that the architecture of the LMA regression network is flexible. We adopt a 2D ResNet~\cite{he2016deep} in all our experiments. \\

\noindent \textbf{Joint loss optimization.} We jointly optimize the registration-based strain network and the LMA regression network in the training process. The total loss function is the sum of strain loss (in Eq.~\eqref{eq:strain}) and TOS loss (in Eq.~\eqref{eq:LMA}), i.e., $l_{\text{strain}} + l_{\text{TOS}}$.
%\begin{equation}
 %   l = l_{\text{strain}} + l_{\text{TOS}}.
%\end{equation}
% \begin{figure*}[btp]
%     \centering
%     \includegraphics[width=\linewidth]{Figures/3d-activation-map-and-boxplot.pdf}
%     \caption{(Left) Boxplot showing TOS regression error and LMA sector classification accuracy 
%  across different methods; (Right) 3D activation maps of 3 subjects, where the regions with severe LMA are presented in red. DENSE results are used for reference in both figures.}
%     \label{fig:boxplot-and-3d-map}
% \end{figure*}
\section{Experiments}
We validate our method on cine CMR images paired with DENSE. A comparison of our multimodal joint learning model with existing deep learning methods, including cine FT based on deformable image registration~\cite{zhang2015finite} and DENSE for LMA detection~\cite{xing2021deep}, is performed.

\noindent \textbf{Data acquisition.} All short-axis cine bSSFP images were acquired during repeated breath holds covering the LV (temporal resolution, $30$-$55$ ms). Cine DENSE was performed in $4$ short-axis planes at basal, two mid-ventricular, and apical levels (with temporal resolution of $17$ ms, pixel size of $2.65^2 \text{ mm}^2$, and slice thickness=$8$mm). Other parameters included displacement encoding frequency $k_e = 0.1 \text{ cycles/mm}$, flip angle $15^{\circ}$, and echo time $= 1.08 \text{ ms}$. All cine and DENSE images are cropped to the size of $128^2$, with $T=40$ time frames for cine and $T=20$ for DENSE.  All LV myocardium segmentation and ground-truth TOS curves were manually labeled by experts.

\noindent \textbf{Experimental settings}. In our experiments, we utilized $118$ left ventricle MRI scan slices from $24$ subjects, divided into $66$ slices for training, $26$ for validation, and $26$ for testing from different subsets of subjects. We first compare our multimodal joint learning model with the baseline algorithms, including cine FT and DENSE-strain for LMA detection and using TOS ($N=128$) mean square error as the evaluation metric. The TOS error (the MSE between predicted TOS from all methods vs. ground truth) is used a evaluation metric. We also employ a second evaluation metric, which is LMA sector classification accuracy. More specifically, we classify sector as LMA region if its TOS value is greater than a specified variable. While any region where the LV myocardium does not start contraction at the first frame (i.e., TOS=$17$ms) is considered as LMA, we take the LMA threshold as $18$ms to avoid small numerical perturbations in all experiments.  
%{\color{blue}The LMA region classification accuracy is also compared, where the LMA region labels are generated from the TOS curves (regions with TOS $\geq18$ms are considered as LMA region).}

We visualize 3D activation maps reconstructed from the TOS prediction. Using myocardium segmentation from sparsely scanned CMR slices, we first construct coordinates for densely sampled points on the myocardium surface through spatial interpolation. A similar interpolation strategy is then used to estimate TOS at those sampled points.

% {\color{red} add details of how the reconstruction is done!!!}

All experiments were trained on an Nvidia 2080Ti GPU using an Adam optimizer. The hyper-parameters are tuned with grid search strategy, and the optimal values are $\sigma=0.03$, $\lambda=\mu=\gamma=0.0001$, $\alpha=1000$ and $\beta=0.005$.

\begin{figure}[!h]
    \centering
    \includegraphics[width=0.95\linewidth]{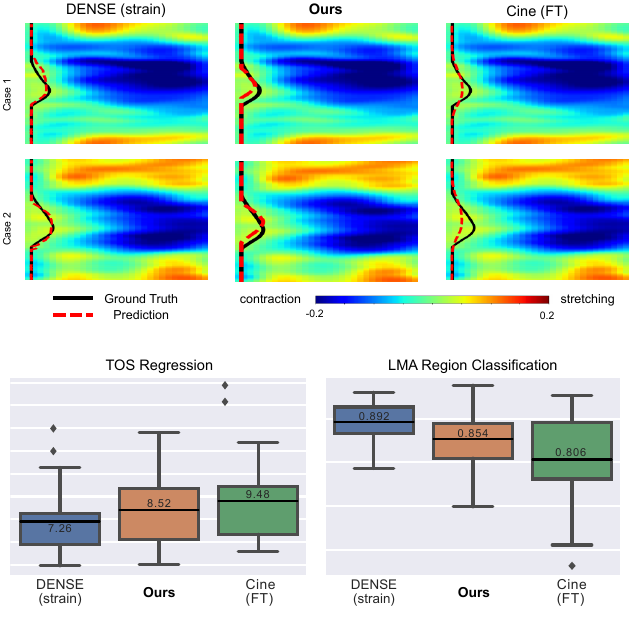}
    \caption{Top to bottom panel: a comparison of TOS prediction from all methods vs. manually labeled TOS (marked in solid black) overlaid on strain matrix; TOS regression mean square error vs. LMA classification accuracy from all methods.}
    \vspace{-1.2em}
    \label{fig:strainmat-and-boxplot}
\end{figure}

\noindent \textbf{Experimental results.} The top panel of Fig.~\ref{fig:strainmat-and-boxplot} shows examples of estimated TOS by our method and all baselines. It shows that DENSE-strain predicted TOS fits the ground truth better than cine FT, especially in the peak region of TOS. Our method is able to bridge the gap between DENSE and cine FT, reaching closer TOS prediction to DENSE. The bottom panel of Fig.~\ref{fig:strainmat-and-boxplot} displays quantitative results of TOS error and LMA region classification error of all methods. Similarly, our method achieves closer accuracy to DENSE with substantially improved performance over cine FT. 

\begin{figure}[!h]
    \centering
    \includegraphics[width=0.9\linewidth]{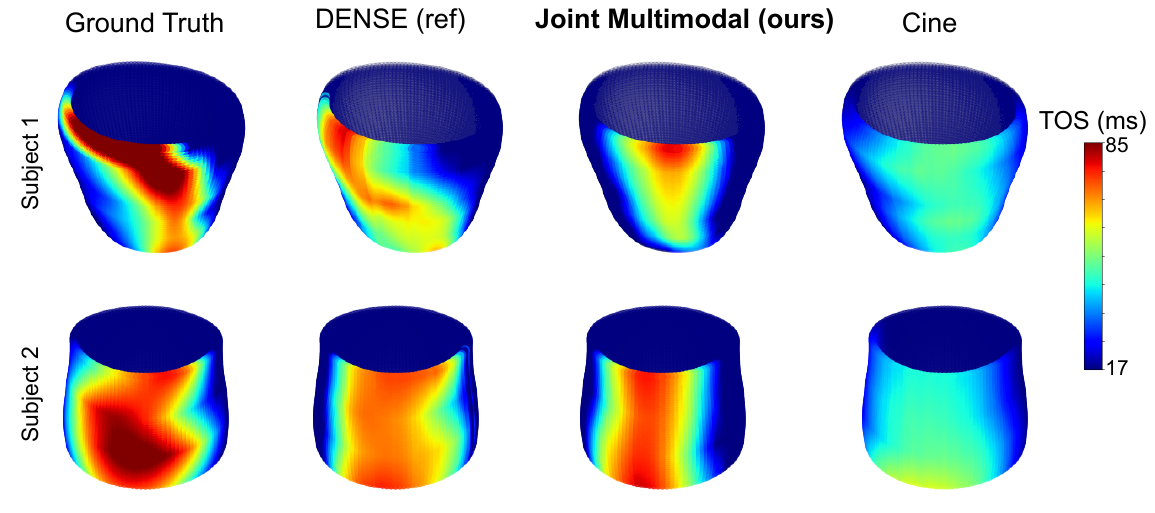}
    \caption{Left to right: a comparison of 3D Activation Maps from ground truth vs. reconstructed from all methods.}
    \label{fig:3d-map}
    \vspace{-1.0em}
\end{figure}

Fig.~\ref{fig:3d-map} shows a comparison of reconstructed 3D activation maps using all methods vs. the ground-truth TOS data. Note that regions with TOS values much larger than $18$ms (shown in red) indicate severe late activation, and normal regions (shown in blue) are typically with small TOS values. Our approaches provide the more accurate LMA region estimation than cine FT.

%Both the regression and multi-task networks are trained on an Nvidia 2080Ti GPU with $11$ GB RAM over maximum $1000$ epochs using an Adam optimizer~\cite{kingma2014adam} with early stop. The hyper-parameters are tuned with grid search strategy, and the optimal values are $\sigma=0.03$, $\lambda_{\text{reg}}=\lambda_{\text{strain}}=w_{\text{LMA}}=0.0001$, $\lambda_{\text{reg}}=1000$ and $\alpha_s=1$. \\

% \setlength{\abovecaptionskip}{10em}
% \begin{figure}[!htb]
%     \centering
%     \includegraphics[width=1.0\linewidth]{Figures/boxplot-SVD.pdf}
%     \vspace{-0.5em}
%     \caption{Boxplot showing TOS regression error and LMA sector classification accuracy cross different methods, where the DENSE result is used for reference.}
%     \label{fig:exp-boxplot}
%     \vspace{-0.5em}
% \end{figure}

\section{Conclusion}
This paper presents a multimodal deep learning framework that provides improved cardiac LMA detection accuracy from routinely acquired standard cine CMR images. Experimental results on LMA detection tasks and 3D activation map visualization show that our work substantially outperforms current approaches based on cine FT, and offers performance that aligns more closely with the achievements with DENSE. Experimental findings in this paper indicate a promising convergence of accessibility and accuracy in the analysis of CMR strain imaging. Our future work will focus on (i) further improve the model accuracy to match the DENSE performance; and (ii) thoroughly validate the model's generalizability as more patient data becomes available. \\
\noindent \textbf{Compliance with ethical standards.} This work was supported by NIH 1R21EB032597. All studies involving human subjects and waiver of consent were approved by our institutional review board. 
\bibliographystyle{IEEEbib-apa}
\bibliography{refs}
% \bibliography{refs-apa}

\end{document}